\documentclass{edm_article}
\usepackage{graphicx}
\usepackage{subcaption}
\usepackage{multirow}
\usepackage{algorithm}
\usepackage{algorithmic}
\usepackage{amsmath}
\usepackage{dsfont}
\usepackage{mathrsfs}

\DeclareMathOperator*{\argmin}{argmin}
\DeclareMathOperator*{\argmax}{argmax}

\begin{document}
\emergencystretch 3em

\title{A Generalized Apprenticeship Learning Framework for Modeling Heterogeneous Student Pedagogical Strategies}

\numberofauthors{5}
\author{
Md Mirajul Islam$^1$, Xi Yang$^2$, John Hostetter$^1$, Adittya Soukarjya Saha$^1$ and Min Chi$^1$\\
        \affaddr{$^1$ North Carolina State University}\\
        \affaddr{$^2$ IBM Research}\\
        \email{mislam22@ncsu.edu, xi.yang@ibm.com, \{jwhostet, asaha4, mchi\}@ncsu.edu}\\
}

\maketitle




\begin{abstract}
A key challenge in e-learning environments like Intelligent Tutoring Systems (ITSs) is to induce effective pedagogical policies efficiently. While Deep Reinforcement Learning (DRL) often suffers from \textbf{\emph{sample inefficiency}} and \textbf{\emph{reward function}} design difficulty, Apprenticeship Learning (AL) algorithms can overcome them. However, most AL algorithms can not handle heterogeneity as they assume all demonstrations are generated with a homogeneous policy driven by a single reward function. Still, some AL algorithms which consider heterogeneity, often can not generalize to large continuous state space and only work with discrete states. In this paper, we propose an expectation-maximization(EM)-EDM, a general AL framework to induce effective pedagogical policies from given optimal or near-optimal demonstrations, which are assumed to be driven by heterogeneous reward functions. We compare the effectiveness of the policies induced by our proposed EM-EDM against four AL-based baselines and two policies induced by DRL on two different but related tasks that involve pedagogical action prediction. Our overall results showed that, for both tasks, EM-EDM outperforms the four AL baselines across all performance metrics and the two DRL baselines. This suggests that EM-EDM can effectively model complex student pedagogical decision-making processes through the ability to manage a large, continuous state space and adapt to handle diverse and heterogeneous reward functions with very few given demonstrations.


\end{abstract}





\section{Introduction}

Reinforcement Learning (RL) and Deep RL (DRL) have experienced significant development in recent years and have been successfully applied in e-learning systems such as intelligent tutoring systems (ITSs) \cite{abdelshiheed2023bridging, chi2011empirically, chi2011evaluation, hostetter2023self, iglesias2009learning, shen2016reinforcement}. More specifically, in an ITS, the pedagogical agent leverages RL to learn a decision-making policy to maximize the expected cumulative \emph{rewards} over an extended period, with the overarching aim of benefiting students in the long run \cite{abdelshiheed2023leveraging, hostetter2023leveraging, ju2022student}. Despite DRL's great success, several obstacles prevent the broader application of DRL to educational systems in practice. One is \textbf{\emph{sample inefficiency}}. For example, it takes Deep Q-Networks (DQN), one type of classic DRL algorithm, hundreds of millions of interactions with the environment to learn a good policy and generalize to unseen states. 
In many previous studies on RL and DRL in ITSs, a prevalent practice involves generating an exploratory corpus. This is typically done by training a cohort of students on an ITS that produces random, yet \emph{reasonable}, decisions. Subsequently, RL is applied to derive pedagogical policies from this training corpus. However, it is noteworthy that despite this approach, a significant portion of prior research aims to derive pedagogical policies from fewer than 3000 logs of student-tutor interactions; the resultant RL policies may sometimes be proven ineffective \cite{DBLP:conf/aied/AusinMBC20, shen2016reinforcement}.

 Another major obstacle is  \textbf{\emph{reward function}} design.  As a fundamental element in RL, the \emph{reward function} defines the goals to achieve when the agent interacts with the environment, which provides the agent incentive to adopt better decision-making behaviors \cite{kaelbling1996reinforcement}. Specifically, the reward function maps each perceived state (or state-action pair) of the environment to a scalar value and gives it as a credit or punishment to the RL agent, and the agent will learn a decision-making policy accordingly to maximize the expected cumulative rewards. Just as supervised models depend heavily on accurate \emph{labels} for the training dataset, the effectiveness and robustness of RL approaches depend heavily on an accurate \emph{reward} function. 
Despite the importance of the reward function in RL, it is usually difficult to design, especially for \emph{human-centric} tasks like education, where multiple factors need to be covered and traded off. Generally, the reward function is hand-crafted beforehand, separate from the policy induction. However, manually designing an appropriate reward function is always labor-intensive and time-consuming \cite{goyal2019using}. It commonly depends on the domain knowledge, and it is hard to avoid the expertise blind spots \cite{abbeel2004apprenticeship}. Moreover, the manually specified reward function will likely be miss-specified, which can be inconsistent with the expected policy \cite{amodei2016concrete}.

To overcome the two obstacles, \textbf{Apprenticeship Learning (AL)} algorithms have been proposed. Instead of inducing the policy under the guidance of a pre-defined reward function and using a large amount of training data, AL aims at learning via observing and imitating a few demonstrations provided by expert agents, who make decisions \emph{optimally} or \emph{near-optimally} with respect to an unknown underlying reward function \cite{abbeel2004apprenticeship}. By learning from demonstrations, the learner agent, \textit{i.e.}, the apprentice, aims to learn a decision-making policy to behave as well as the experts. Getting rid of the difficulties in reward function design in RL, AL has been widely applied in various applications \cite{asoh2013application, hostetter2023xai, pan2019dissecting,rafferty2015inferring,wang2020inferring}. 

AL approaches have been categorized as being either online or offline. In the former category, the agent learns while interacting with the environment; the latter learns the policy from pre-collected data.
The existing AL are commonly \emph{online}, requiring interacting with the environment iteratively for collecting new data and then updating the model accordingly \cite{abbeel2004apprenticeship, ziebart2008maximum, finn2016guided}.  Online AL algorithms are generally appropriate for domains where interacting with simulations and actual environments is computationally cheap and feasible. On the other hand, for domains such as e-learning, building accurate simulations or simulated students is incredibly challenging because human learning is a rather complex, poorly understood process; moreover, learning policies while interacting with students can not only be unethical but not allowed. Therefore, we focus on offline AL approaches. A recently proposed energy-based distribution matching (EDM) approach \cite{jarrett2020strictly} has advanced the state-of-the-art in \emph{offline} AL. However, EDM assumes all demonstrations are generated using a homogeneous policy with a \emph{single} reward function. In real-world scenarios, the reward function can be \textbf{heterogeneous} across the demonstrations. For example, in the education domain, when students make decisions during learning, their reward functions can be learning-oriented, efficient-oriented, or not learning \cite{yang2020student}. Therefore, in this work, to handle the heterogeneous reward functions, we propose an expectation-maximization(EM)-EDM, a general AL framework \emph{to induce pedagogical policies from given optimal or near-optimal demonstrations, which are assumed to be driven by multiple heterogeneous reward functions}.

While EM-Inverse Reinforcement Learning (EM-IRL) was introduced in \cite{yang2020student}, it is limited because it can only handle relatively small discrete state representation (17 discrete features involved in \cite{yang2020student}). In contrast, our proposed EM-EDM can handle over 140 continuous features.
In this study, we focused on the decisions on whether to present the next problem as a Worked Example (WE), a Problem Solving (PS), or a Faded Worked Example (FWE). In WE, students were given a detailed example showing how the tutor solves a problem; in PS, by contrast, students were tasked with solving the same problem on their own on the ITS; in FWEs, the students and the tutor \emph{co-construct} in that their solutions are intertwined.

In AL, it is usually assumed that the demonstrations are optimally or near optimally executed by experts \cite{abbeel2004apprenticeship}. Thus, the quality of the demonstrations matters in order to induce more effective policies. In this study, rather than gathering exploratory data in DRL, our EM-EDM will derive pedagogical policies directly from the pedagogical decisions and behaviors demonstrated by \textbf{\emph{students}} while interacting with the ITS. In this study, we involve 53 ``optimal" or ``near-optimal" demonstrations, which empower students to make their own pedagogical decisions during learning, leading to positive learning outcomes (see details in Section~\ref{TrajSelection}).

We compare the effectiveness of the policies induced by our proposed EM-EDM against four AL-based baselines using the 53 demonstrations and two policies induced by DRL using 2,716 and 1,819 student-ITSs interactive trajectories, respectively. This evaluation is conducted on two tasks that are related but distinct from each other. For Task 1, we adhere to standard AL research practices by conducting training and evaluation of the methods for pedagogical action prediction. This is accomplished through 5-fold cross-validation, utilizing all selected demonstrations.
Task 2 holds greater relevance and practical significance in the educational data mining field. It involves using trajectories from a previous semester as training data and trajectories from a subsequent semester as testing data. The objective is to assess whether the clustering results obtained from the training data of the previous semester can effectively predict students' actions in the later semester. 
Our overall results showed that, for both tasks, EM-EDM outperforms the four AL baselines across all performance metrics as well as the two DRL baselines. This suggests that EM-EDM is more data efficient and effective in learning students' heterogeneous strategies in ITSs. 

In summary, our contributions can be outlined as follows:
\begin{enumerate}
\item  The key contribution lies in the capability of EM-EDM to effectively model complex student pedagogical decision-making processes through the ability to manage a large, continuous state space and its adaptability in handling diverse and heterogeneous reward functions.
\item  The empirical evaluation of EM-EDM involves comparing four state-of-the-art AL baselines and two DRL baselines on two significant tasks. 
\item  The results obtained from EM-EDM showed that students' demonstrations exhibit heterogeneous reward functions. Notably, EM-EDM can induce distinct pedagogical policies for different reward functions even with as few as 24 demonstrations. 
\end{enumerate}

\section{Related Work}

\subsection{RL and DRL for Pedagogical Policy Induction}

In general, there have been two types of prior research on using Reinforcement Learning (RL) to educational policy induction: classic RL and DRL approaches. With the ultimate goal of helping students in the long run, the pedagogical agent in an ITS uses RL to learn a decision-making policy that maximizes the predicted cumulative benefits over a prolonged period \cite{ju2022student}. Despite DRL's enormous success, several challenges are preventing its wider implementation in actual educational institutions; sample inefficiency is one. For instance, learning a suitable policy and generalizing to unknown states requires hundreds of millions of interactions with the environment for Deep Q-Networks (DQN), one kind of classic DRL method. 

The design of reward functions is another significant barrier. They are a key component of reinforcement learning and establish objectives for the agent to meet when interacting with the environment. This motivates the agent to make better decisions \cite{kaelbling1996reinforcement}. In particular, the reward function assigns a credit or punishment to each perceived state (or state-action combination) of the environment. Based on this, the RL agent learns a decision-making policy that maximizes the expected cumulative rewards. An accurate reward function is critical to the efficacy and robustness of reinforcement learning techniques, much as supervised models rely significantly on accurate labels for the training dataset. 

Even though the reward function is crucial to reinforcement learning, designing one is typically challenging, particularly for tasks that focus on people, like teaching, and require balancing various considerations. Usually, the policy induction is done separately from the reward function, which is manually created previously. However, creating a suitable reward function by hand is usually time-consuming and labor-intensive \cite{goyal2019using}. 

To summarize, previous research indicates that RL and DRL-induced pedagogical policies can be sample inefficient and require handling the challenge of reward function design.

\subsection{Apprenticeship Learning (AL)}
Behavior cloning \cite{syed2012imitation,raza2012teaching} is a traditional offline AL technique that learns a mapping from states to actions by avariciously copying the best practices of experts who have been shown \cite{pomerleau1991efficient}. Several inverse reinforcement learning (IRL)-based  \cite{abbeel2004apprenticeship,ziebart2008maximum} and adversarial imitation learning-based \cite{ho2016generative} techniques have been developed to better capture the data distribution in experts' presentations.

In general, iterative loops are used in IRL-based approaches to 1) infer a reward function, 2) induce a policy using traditional reinforcement learning, 3) roll out the learned policy, and 4) update the reward parameters in response to discrepancies between the roll-out behaviors and expert demonstrations. The technical and theoretical drawbacks of the original IRL approaches that adhere to the aforementioned online model will be retained when applied to an offline environment. Furthermore, the reward is typically modeled by IRL-based approaches using a certain tractable format, such as a linear function that maps states and state-action pairings to reward values \cite{abbeel2004apprenticeship,ziebart2008maximum,babes2011apprenticeship}. Moreover, certain batch-IRL have been proposed in order to prevent the learned policy from being implemented.

Typically, adversarial imitation learning-based techniques involve learning a discriminator to discern learned behaviors from expert demonstrations and a generator to implement the policy iteratively. Some off-policy learning techniques based on off-policy actor-critic have been developed under the batch setting in order to avoid rolling out the policy \cite{kostrikov2018discriminator,kostrikov2019imitation}. However, according to \cite{ho2016generative}, these techniques inherit the complicated alternating max-min optimization from general adversarial imitation learning.

Jarrett \textit{et al.} proposed and evaluated the \textbf{EDM} \cite{jarrett2020strictly} on a wide range of benchmarks, online environments like \textit{e.g.}, Acrobot, LunarLander, and BeamRider, as well as offline environment MIMIC-III. It was shown that EDM can outperform both \emph{IRL-based} and \emph{adversarial imitation learning-based} methods. In this work, EDM serves as one of our baselines.

\subsection{AL with Multiple Intentions}
A few AL methods have been proposed to consider multiple reward functions. A Bayesian multi-task IRL was proposed, which models the heterogeneity of reward functions by formalizing it as a statistical preference elicitation via a joint reward-policy prior \cite{dimitrakakis2011bayesian}. Choi and Kim integrated a Dirichlet process mixture model into Bayesian IRL to cluster the demonstrations \cite{choi2012nonparametric}. Using a Bayesian model, they incorporated the domain knowledge of multiple reward functions. Similarly, Arora \textit{et al.} combined the Dirichlet process with a maximum entropy IRL to learn the clusters of demonstration with different reward functions \cite{arora2021min}. 

Babes \textit{et al.} derived an EM-based IRL approach that clusters trajectories based on their different reward functions \cite{babes2011apprenticeship}. As a component of the EM-based IRL, a maximum-likelihood IRL uses a gradient ascent method to optimize the reward parameters, which has successfully identified unknown reward functions. Xi et al. used EM-Inverse Reinforcement Learning (EM-IRL) to subtype students from their pedagogical behavior data \cite{yang2020student}. Still, as most other multiple intentions works, it is limited because it can only handle discrete state representation (17 discrete features involved in \cite{yang2020student}) and can not generalize to large continuous state space.

\subsection{WE, PS, and FWE}

Many studies have examined the effectiveness of WE, PS, and FWE, as well as their different combinations \cite{van2011effects,renkl2002,najar2014}. 
Renkl et al. \cite{renkl2002} compared WE-FWE-PS with WE-PS pairs, and the results showed that the WE-FWE-PS condition significantly outperformed the WE-PS condition on posttest scores. Similarly, Najar et al. \cite{najar2014} compared adaptive WE/FWE/PS with WE-PS pairs and found that the former is significantly more effective than the latter on improving student learning. 
Overall, it is demonstrated that adaptively alternating amongst WE, PS, and FWE is more effective than hand-coded expert rules in improving student learning. 
However, when students make decisions among WE, PS, and FWE, there's no significant difference with tutors making decisions on students' learning performance \cite{guojing2017tutorstudent}. As far as we know, no prior research has explored how to combine students' decision-making with RL-induced policy decision-making to facilitate learning.

\section{Methods}
\subsection{Our Methodology: EM-EDM }
\vspace{2pt}

AL generally follows the procedure of 1) taking the demonstrations as input to infer the latent reward function, based on which a policy can be induced, then 2) rolling out the learned policy to update the reward function by minimizing the divergence between the rolled-out behaviors versus the demonstrations. This procedure will be conducted iteratively until convergence. 

\begin{figure}
    \centering
    \includegraphics[scale=0.4]{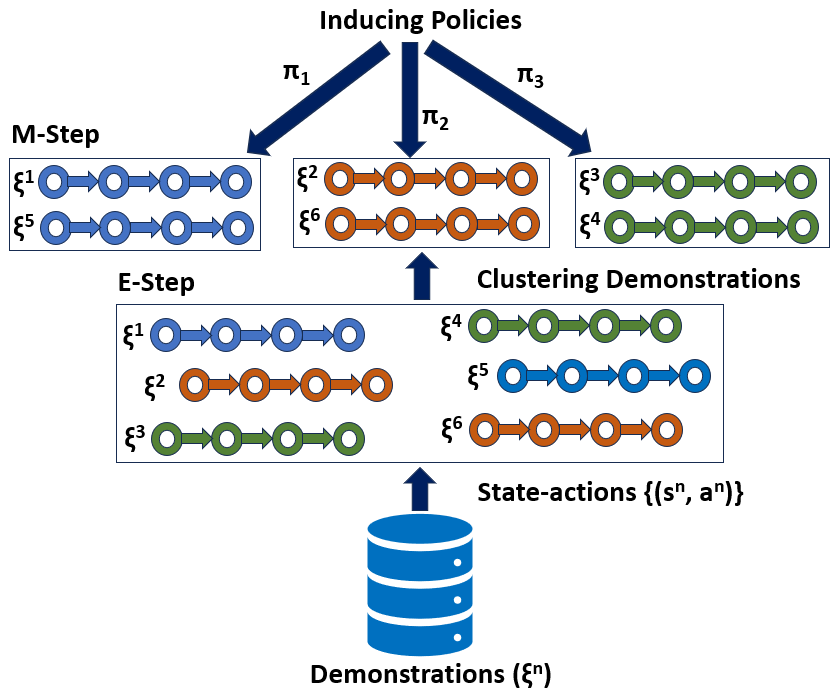}
    \caption{EM-EDM framework}
    \label{fig:emedm}
    \Description{EM-EDM framework}
\end{figure}

The input of EM-EDM contains $N$ demonstrated trajectories $\{\xi^{n}\} = \{(\mathbf{s}_t^n, a_{t}^{n})|t=1, ..., T^{n}; n = 1,..., N\}$ provided by experts, where $\mathbf{s}_t^n\in \mathbb{R}^m$ is the $t$-th multivariate state with $m$ features, and $a_{t}^{n}$ is the corresponding action in the $n$-th trajectory with the length of $T^{n}$. In this work, we assume that $\{\xi^{n}\}$ are carried out with \emph{multiple} policies. EM-EDM would cluster the trajectories first and then induce policies for each cluster to induce effective policies. Taking our ITSs as an example, EM-EDM will cluster student trajectories and induce different policies or strategies over learned clusters (as shown in Figure \ref{fig:emedm}).

\begin{algorithm}
    \caption{\textbf{EM-EDM}}
    \label{alg:emedm}
    \renewcommand{\thealgorithm}{}
    \begin{algorithmic}[1]
           \STATE \textbf{Input}: Expert trajectories $\{\xi^{n}\}$, number of clusters $\{K\}$\\ 
           \STATE \textbf{Initialize}: Prior probability $\nu_{j}$ and policy parameter $\theta_{j}$, $j = 1,...K$ randomly
            \REPEAT
                \STATE \textbf{E Step}: Compute the $u_{ij}$, $i = 1,...N$, $j = 1,...K$
                \STATE \textbf{M Step}: Update $\nu_{j}$ and learn $\theta_{j}$ via EDM, $j = 1,...K$
            \UNTIL{\textit{stop\_criteria} is True}
            \STATE \textit{Output}: Cluster-wise policies $\{\Pi^{\theta}_j|j=1,...K\}$ 
   \end{algorithmic}
\end{algorithm}

\subsubsection{EDM}

As a strictly \emph{offline} AL, energy-based distribution matching (EDM) \cite{jarrett2020strictly} can learn the policy merely based on the experts' demonstrations, not requiring any knowledge of model transitions or off-policy evaluations. It assumes that the demonstrations $\{\xi^{n}\}$ are carried out with a policy $\Pi^{\theta}$ parameterized by $\theta$, driven by a \emph{single} reward function. 

We denote the state-action pair as $(\mathbf{s}, a)$. The occupancy measures for the demonstrations and the learned policy are denoted as $\rho_{\xi}$ and $\rho_{\Pi^{\theta}}$, respectively. 
The probability density for each state-action pair can be measured as:  $\rho_{\Pi^{\theta}}(\mathbf{s},a) =  \mathbb{E}_{\Pi^{\theta}}[\sum_{t=0}^{\infty} \gamma^{t} \mathds{1}{\{\mathbf{s}_{t}=\mathbf{s}, a_{t}=a\}}]$, where $\gamma$ is a discount factor, then the probability density for each state can be measured by: $\rho_{\Pi^{\theta}}(\mathbf{s}) = \sum_{a} \rho_{\Pi^{\theta}}(\mathbf{s},a)$. To induce the policy $\Pi^{\theta}$, our goal is minimizing the KL divergence between $\rho_{\xi}$ and $\rho_{\theta}$: 

\vspace{-0.2cm}
\begin{equation} 
    \argmin_{\theta} D_{KL}(\rho_{\xi}||\rho_{\theta}) = \argmin_{\theta} -\mathbb{E}_{\mathbf{s},a \sim \rho_{\xi}} \log \rho_{\theta}(\mathbf{s},a)
\end{equation}

Since $\Pi^{\theta}(a|\mathbf{s}) = \rho_{\Pi^{\theta}}(\mathbf{s},a)/\rho_{\Pi^{\theta}}(\mathbf{s})$, we can formulate the objective function as: 

\vspace{-0.2cm}
\begin{equation}
\label{eq:6}
    \argmin_{\theta} - \mathbb{E}_{\mathbf{s} \sim \rho_{\xi}} \log \rho_{\Pi^{\theta}}(\mathbf{s}) - \mathbb{E}_{\mathbf{s},a \sim \rho_{\xi}} \log \Pi^{\theta}(a|\mathbf{s}) 
\end{equation} 

\noindent When there is no access to roll out the policy $\Pi^{\theta}$ in an \emph{online} manner, $\rho_{\Pi^{\theta}}(\mathbf{s})$ in the first term of Eq.\eqref{eq:6} would be challenging to estimate. EDM can handle this issue by utilizing an energy-based model \cite{grathwohl2019your}.

According to energy-based model, the probability density $\rho_{\Pi^{\theta}}(\mathbf{s}) \propto e^{-E(\mathbf{s})}$, with $E(\mathbf{s})$ being an energy function. 
Then the occupancy measure for state-action pairs can be represented as:  $\rho_{\Pi^{\theta}}(\mathbf{s},a) = e^{f_{\Pi^{\theta}}(\mathbf{s}) [a]} / Z_{\Pi^{\theta}}$, and the occupancy measure for states can be obtained by marginalizing out the actions: $\rho_{\Pi^{\theta}}(\mathbf{s}) = \sum_{a} e^{f_{\Pi^{\theta}}(\mathbf{s}) [a]} / Z_{\Pi^{\theta}}$. Herein, $Z_{\Pi^{\theta}}$ is a partition function, and $f_{\Pi^{\theta}}: \mathbb{R}^{|S|} \rightarrow \mathbb{R}^{\mathbb{|A|}}$ is a parametric function that mapping each state to $A$ real-valued numbers. 


The parameterization of $\Pi^{\theta}$ implicitly defines an energy-based model over the state's distribution, where the energy function can be defined as: $E_{\Pi^{\theta}}(\mathbf{s}) = -\log \sum_{a} e^{f_{\Pi\theta}(\mathbf{s})[a]}$. 
Under the scope of the energy-based model, the first term in Eq.\eqref{eq:6} can be reformulated as an ``occupancy” loss: 

\vspace{-0.2cm}
\begin{equation}
\label{eq:8}
    \mathcal{L}_{\rho}(\theta) = \mathbb{E}_{\mathbf{s} \sim \rho_{\xi}} E_\theta (\mathbf{s}) - \mathbb{E}_{\mathbf{s} \sim \rho_{\Pi^{\theta}}} E_\theta (\mathbf{s}) 
\end{equation}

\noindent where $\nabla_\theta \mathcal{L}_\rho (\theta) = - \mathbb{E}_{\mathbf{s} \sim \rho_{\xi}} \nabla_\theta \log \rho_{\Pi^{\theta}}(\mathbf{s})$ can be solved by existing optimizers, \textit{e.g.}, stochastic gradient Langevin dynamics \cite{welling2011bayesian}. Therefore, by substituting the first term in Eq.\eqref{eq:6} as Eq.\eqref{eq:8} via energy-based model, we can derive a \emph{surrogate objective function} to get the optimal solution without the need of \emph{online} rolling out the policy. 


\subsubsection{EM-EDM}

To deal with \emph{multiple} reward functions varying across the demonstrations, Babes-Vroman \textit{et al.} proposed an EM-based inverse reinforcement learning \cite{babes2011apprenticeship} by iteratively clustering the demonstrations in \textit{E-step} and inducing policies for each cluster by IRL in \textit{M-step}. Specifically, in the \textit{M-step}, they explored several IRL methods based on the \emph{discrete} states, which is not scalable for large continuous state spaces, \textit{e.g.}, ITSs logs. 
Enlightened by this EM framework and the success of EDM, we proposed an EM-EDM (Algorithm \ref{alg:emedm}) \cite{xi2023themes}. 

Taking trajectories $\{\xi^{i} | i=1,..., N\}$ as input, with $N$ being the number of trajectories, EM-EDM aims to cluster these trajectories and learn the cluster-wise policies $\{\Pi_j|j=1,...K\}$, where $K$ is the number of clusters. $\nu_{j}$ and $\theta_{j}$, denote the prior probability and the policy parameter for each cluster, and both are randomly initialized.  
The objective function of EM-EDM is to maximize the log-likelihood defined in Eq.\eqref{eq:10}:


\begin{equation}
\label{eq:10}
    \argmax_{\theta_j} \underset{j=1}{\overset{K}{\sum}} \underset{i=1}{\overset{N}{\sum}} log(u_{ij})
\end{equation}

\noindent where $u_{ij}$ denotes the probability that trajectory $\xi^{i}$ follows the policy of the $j$-th cluster. It is defined in Eq.\eqref{eq:5}, with $U$ being a normalization factor.

\begin{equation}
\label{eq:5}
     u_{ij} = Pr(\xi^{i}|\theta_{j}) = \underset{(\mathbf{s},a) \in \xi^{i}}{\prod} \frac{\Pi_{\theta_{j}}(\mathbf{s},a)\nu_{j}}{U}
\end{equation}

During the EM process, in the \textit{E-step}, the probability that trajectory $\xi^{i}$ belonging to cluster $j$ is calculated by Eq.\eqref{eq:5}. 
Then in the \textit{M-step}, the prior probabilities are updated via $\nu_{j} = \sum_{i} u_{ij}/N$, and the policy parameters $\theta_{j}$ is learned by EDM. The \textit{E-step} and \textit{M-step} are iteratively executed until converged. Finally, the output of EM-EDM is the clustered trajectories with their respective policies.

\subsection{Apprenticeship Learning Baselines}

\subsubsection{Behavior cloning (BC)}
Behavior cloning \cite{syed2012imitation,raza2012teaching} is a traditional offline AL technique that learns a mapping from states to actions by avariciously copying the best practices of experts who have been shown \cite{pomerleau1991efficient}. It is a supervised learning approach used in AL where an agent learns to imitate an expert by directly mapping observations to actions based on the expert's demonstrations. The objective is to mimic the expert's behavior. The learning process involves minimizing the difference between the agent's actions and those demonstrated by the expert \cite{bain1999bc}.

In a simplified form, let's consider a dataset of expert demonstrations where each example consists of an observation-action pair: $(s_i, a_i)$, where $s_i$ is an observation and $a_i$ is the corresponding action. The behavior cloning objective is to learn a policy $\pi_{\theta}(a|s)$ parameterized by $\theta$ that approximates the expert's behavior.

The loss function for behavior cloning can be defined as the standard supervised learning loss:

\begin{equation}
    L(\theta) = \frac{1}{N} \sum_{i=1}^{N} \lVert \pi_{\theta}(a_i|s_i) - a_i \rVert^2    
\end{equation}

Where $N$ is the length of trajectories. The goal is to minimize this loss, and the optimal $\theta$ corresponds to a policy that imitates the expert's actions.

It's important to note that while BC is straightforward, it can be sensitive to distribution shifts and compounding errors, especially when the learned policy deviates from the expert's demonstrations.

\subsubsection{Generative Adversarial Imitation Learning (GAIL)}
Generative Adversarial Imitation Learning (GAIL) is another AL approach that uses a generative adversarial network (GAN) \cite{goodfellow2014generative} to model the reward function, making the learned policy less sensitive to differences between the demonstrator and the learner \cite{ho2016gail}. GAIL aims to generate a reward function that encourages the learner's policy to match the demonstrated behavior.

Given a set of expert trajectories $\tau_E \sim \pi_E$, and initial policy and discriminator parameter to be $\theta_0, w_0$, GAIL iteratively updates the discriminator parameters from $w_i$ to $w_{i+1}$ with the gradient as per Equation \ref{eqn:gail1}.

\begin{equation}
    \hat{E}_{\tau_i} \left[ \nabla_w \log(D_w(s, a)) \right] + \hat{E}_{\tau_E} \left[ \nabla_w \log(1 - D_w(s, a)) \right]
    \label{eqn:gail1}
\end{equation}

And then, take a KL-constrained natural gradient step with Equation \ref{eqn:gail2}.

\begin{equation}
\begin{split}
   \hat{E}_{\tau_i} \left[ \nabla_{\theta} \log \pi_{\theta}(a | s) Q(s, a) \right] - \lambda \nabla_{\theta} H(\pi_{\theta}),\\
   \text{ where } Q(\bar{s}, \bar{a}) = \hat{E}_{\tau_i} \left[ \log(D_{w_{i+1}}(s, a)) \,|\, s_0 = \bar{s}, a_0 = \bar{a} \right]
\end{split}
\label{eqn:gail2}
\end{equation}



Where $\pi_{\theta}$ is the policy of the learner, and $\pi_{\text{E}}$ is the expert policy. The discriminator aims to maximize this objective, while the learner's policy minimizes it. The reward function is derived from the discriminator's output.

GAIL addresses distribution shift issues in AL and is more robust than behavior cloning. However, it may face challenges such as mode collapse during GAN training.

\subsubsection{Adversarial Inverse Reinforcement Learning (AIRL)}
Adversarial Inverse Reinforcement Learning (AIRL) is an adversarial approach in AL similar to GAIL that adversarially trains a policy against a discriminator that aims to distinguish the expert demonstrations from the learned policy. Unlike GAIL, AIRL recovers a reward function more generalizable to changes in environment dynamics \cite{fu2018learning}. 

Given a set of expert demonstrations: $\tau_i^E$, initial policy $\pi$ and discriminator $D_{\theta,\phi}$, AIRL iteratively collect trajectories $\tau_i$ by executing $\pi$, train the discriminator via binary logistic regression to classify expert data from samples. The reward is updated in each step following the Equation \ref{eqn:airl1}. Later, the policy is updated with respect to reward using any policy optimization method.

\begin{equation}
r_{\theta,\phi}(s, a, s') = \log D_{\theta,\phi}(s, a, s') - \log(1 - D_{\theta,\phi}(s, a, s'))
\label{eqn:airl1}
\end{equation}

where $\pi_{\theta}$ is the policy of the learner, and $\pi_{\text{expert}}$ is the expert policy. Like GAIL, the discriminator aims to maximize this objective while the learner's policy minimizes it. The reward function is derived from the discriminator's output.

AIRL offers advantages such as handling distribution shifts and being less sensitive to suboptimal demonstrations. However, it may be computationally expensive.

\subsection{Two Deep RL Baselines : CQL \& CQL-T}

\subsubsection{CQL}
A good number of prior research has explored the Deep RL (DRL) (e.g., \cite{ju2020aiedcritical,mandel2014offline,rowe2015improving}), and it has shown that they can be used to induce effective pedagogical policies for ITSs \cite{mandel2014offline,wang2017interactive}. As a DRL baseline, we choose Conservative Q-Learning (CQL) \cite{kumar2020conservative}, known for addressing overestimation issues in comparison to various existing RL methods, including Deep Q-learning, which is extensively explored in prior DRL studies within ITSs.


CQL represents a Q-learning or actor-critic algorithm designed to learn Q-functions, ensuring that the anticipated policy value under the learned Q-function provides a conservative estimate of the actual policy value. To achieve Q-values with lower bounds, CQL goes beyond standard minimization by concurrently minimizing the Q-function under a selected distribution and maximizing it under the data distribution \cite{kumar2020conservative}. The training objective for the Q-function is accordingly structured to attain this goal:

\begin{equation}
\begin{aligned}
\hat{Q}_{\mathrm{CQL}}^\pi:=\arg \min _Q \alpha \cdot(\underbrace{E_{\mathbf{s} \sim \mathcal{D}, \mathbf{a} \sim \mu(\mathbf{a} \mid \mathbf{s})}[Q(\mathbf{s}, \mathbf{a})]}_{\text {minimize Q-values }}\\
-\underbrace{E_{\mathbf{s} \sim \mathcal{D}, \mathbf{a} \sim \hat{\pi}_\beta(\mathbf{a} \mid \mathbf{s})}[Q(\mathbf{s}, \mathbf{a})]}_{\text {maximize Q-values under data }}) \\
+\frac{1}{2} \underbrace{E_{\mathbf{s}, \mathbf{a}, \mathbf{s}^{\prime} \sim \mathcal{D}}\left[\left(Q(s, a)-\left(\mathcal{R}(\mathbf{s}, a)+\gamma\left[\max _{a \in \mathcal{A}} Q\left(\mathbf{s}^{\prime}, a\right)\right]\right)^2\right]\right.}_{\text {standard Bellman error }}
\end{aligned}    
\end{equation}

\vspace{2.5pt}
Then, the optimization of the policy is done w.r.t: \\~\\
\begin{equation}
    \begin{aligned}
\hat{Q}_{\mathrm{CQL}}^\pi: \pi \leftarrow \arg \max _\pi E_\pi\left[\hat{Q}_{\mathrm{CQL}}^\pi\right]
\end{aligned}
\end{equation}

The offline training process continues until it hits convergence, a predetermined error threshold, or the maximum number of iterations. As the amount of accessible data increases, the magnitude of $\alpha$ may decrease, so over here, $\alpha$ > 0 represents a trade-off factor \cite{kumar2020conservative}. Here, $\hat{\pi}_{\beta}$ denotes the behavior policy,
$\gamma \in [0, 1]$ denotes the discount factor, $\mathbf{s}$ is the current state, $\mathbf{s}'$ is the next state and $\mathscr{R}$ corresponds to the reward function.

\subsubsection{CQL-T}
CQL-T refers to the same methodology for policy induction as CQL except that it utilizes a combined reward function of both student training time and Normalized Learning Gain (NLG) motivated by prior research that response time reveals student proficiency \cite{schnipke2002exploring}. In particular, there was a significant negative correlation between student average response time and student final exam score taken at the end of the semester \cite{gonzalez2007innovative}.

\section{Experiment}
\begin{figure}
    \centering
    \includegraphics[scale=0.25]{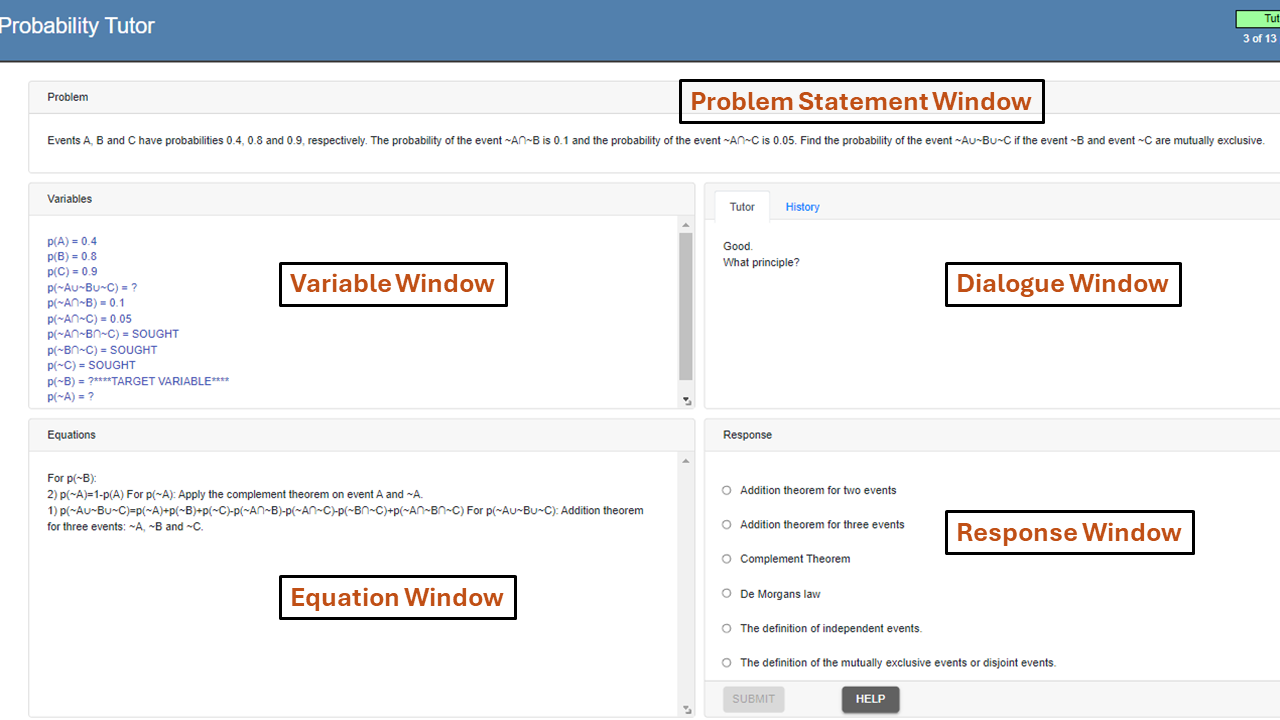}
    \caption{The interface of our probability tutor}
    \label{fig:pyrenees-interface}
    \Description{The interface of our probability tutor}
\end{figure}

We gathered our data by letting undergraduate students from the same major and same year complete an online intelligent tutoring system (ITS) (Figure \ref{fig:pyrenees-interface}) that introduced them to probability concepts like Bayes' Theorem and Addition Theorem. Using training problems, the students were guided through the instruction. The tutor gave detailed instructions, quick responses, and on-demand assistance for every problem. The assistance was given in the form of progressively more detailed hints. The bottom-out hint, the final one in the sequence, gave the students precise instructions. Students could decide throughout training how to solve the next stage pedagogically by either working through it alone, watching the tutor work through it, or doing it collaboratively. The tutor will ask questions to get the answer if they want to solve it independently; if not, the instructor will show or tell.

\subsection{Data Collection}

Our experiment involves 128 students from two semesters, Spring 21 (67 students) and Spring 22 (61 students). The textbook, pre-test, training, and post-test were the four stages every student taking part in our data collection process went through. All students studied the domain fundamentals from a textbook on probability during the textbook. They studied examples of each principle, read a general explanation, and worked through various single- and multi-principle problems. After that, the students completed a pretest with 8 problems. They would not receive feedback on their responses during this phase and would not be permitted to revisit previous questions (this also applied to the post-test). During training, twelve problems are shown in a fixed sequence, allowing students to practice by choosing to solve it alone (PS), solving it \emph{collaboratively} with the ITS (FWE), or seeing the solution as a worked-out example (WE). Students work through the problem step-by-step, define variables, type equations, etc. The fewest steps required to complete each training problem varied between 20 and 50. While making decisions on how to solve the next problem in training, if PS is chosen, the ITS asks questions to elicit the next step's answer from the student; if FWE is chosen, the ITS chooses to elicit or tell the answer; if WE is chosen, the ITS tells the answer. There were between three and eleven domain principles needed to address each problem. At last, every student completed the post-test, which comprised 12 questions. Four problems were more complex, non-isomorphic multiple-principle problems, and the remaining 8 were isomorphic to the problems provided in the pre-test phase.

Students had to write and solve one or more equations to arrive at an answer for the pre and post-tests. Three scoring criteria were applied: binary, partial credit, and one point for each premise. A solution received one point under the binary rubric for being entirely correct and zero points for being incorrect. The percentage of correctly applied principles seen in the solution determined each problem's score under the partial credit rubric. For example, learners would receive a 0.8 score if they correctly implemented 4 of the 5 potential principles. A point was awarded under the one-point-per-principle rubric for every correctly applied principle. Each student's answers to all the pre-test and post-test problems were double-blind, graded by 2 experienced graders, and then aggregated into one final grade by resolving disagreements if there were any. All test scores were normalized to the range of [0 - 100] for comparison purposes.

Moreover, all students' data were obtained anonymously through an exempt IRB-approved protocol, and we scored them using double-blinded grading rubrics. All data were de-identified, and no demographic data or grades were collected. This research seeks to remove societal harms from lower engagement and retention of students who need more personalized interventions for introductory Computer Science Courses.

\subsection{States \& Actions}

\subsubsection{142 Continuous State Space}
142 continuous state features were extracted from the student system interaction log data. Here is a brief description of the features:

\noindent $\bullet$ \textbf{Autonomy(10 features)}: the amount of work done by a student, such as the number of elicits since the last tell;\\
\noindent $\bullet$ \textbf{Temporal (29 features)}: time-related information about the student's behavior, such as the average time per step;\\
\noindent $\bullet$ \textbf{Problem Solving (35 features)}: information about the current problem-solving context, such as problem difficulty;\\
\noindent $\bullet$ \textbf{Performance (57 features)}: information about the student's performance so far, such as the percentage of correct entries;\\
\noindent $\bullet$ \textbf{Hints (11 features)}: information about the student's hint usage, such as the total number of hints requested.

\subsubsection{Three Pedagogical Action Space}
The students can take three decisions which can be interpreted as actions which are elicit(PS), tell(WE) or collaboration(FWE), i.e., to elicit the solution by themselves through asking questions, to let the tutor tell them the solution directly or to work with the tutor collaboratively towards the solution respectively.

\subsection{Selecting Demonstrations (A total of 53)}\label{TrajSelection} 
To enhance the selection of high-quality demonstrations for inducing more accurate yet effective AL policies, it is typically believed that the experts are performing the demonstrations in an optimal or near-optimal manner \cite{abbeel2004apprenticeship}. 
Our original dataset contains 128 students from two semesters, Spring 21 (67 students) and Spring 22 (61 students). Each student spent \~ 2 hours on the probability ITS and completed around 400 steps.
To select higher-quality trajectories from all students' interaction with the ITS, we use a qualitative measurement called Quantized Learning Gain (QLG) \cite{lin2017comparisons}, which is a binary qualitative measurement of students' learning gains from the pretest to the posttest to determine whether a student has benefited from a learning environment \cite{mao2018tracing}.

\begin{figure}
    \centering
    \includegraphics[scale=0.3]{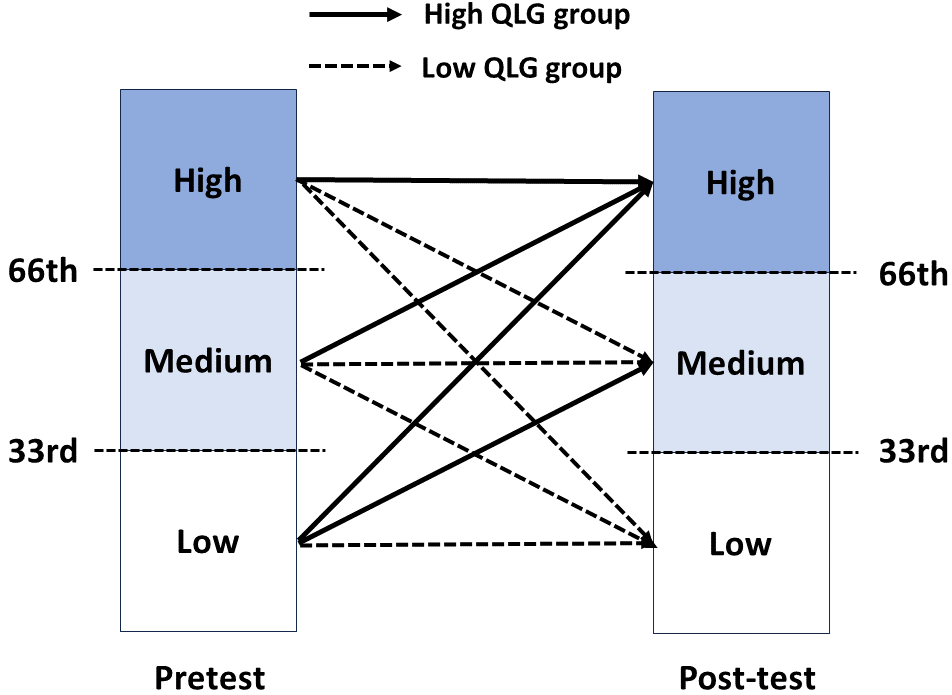}
    \caption{Quantized Learning Gain}
    \label{fig:qlg}
    \Description{Quantized Learning Gain}
\end{figure}

To calculate QLG, first, students were split into low, medium, and high-performance groups based on whether they scored below the 33rd percentile, between the 33rd and 66th percentile, or higher than the 66th percentile in pre-test and post-test, respectively. Once a student's pre-test and post-test performance groups (high, medium, or low) are decided, the student is a ``High'' QLG if he/she moved from a lower performance group to a higher performance group from pre-test to post-test or remained in ``high" performance groups; whereas a ``low" QLG is assigned to the student if he/she either moved from a higher performance group to a lower performance group from pre-test to post-test or stayed at a ``low" or ``medium" groups (as shown in Figure \ref{fig:qlg}). In Figure \ref{fig:qlg}, solid lines represent the formation of the High QLG groups, and dashed lines represent the formation of the Low QLG groups. Following this procedure, Table \ref{tab:qlg_stat} illustrates the number of trajectories in each category across the two semesters. This procedure resulted in a total of \textbf{53} ``high-quality" student trajectories, which are treated as optimal or near-optimal demonstrations in the following analysis.


\renewcommand{\arraystretch}{1.5}
\begin{table}
\footnotesize
    \centering
    \caption{\centering No. of Students in High QLG Group}
    \setlength\tabcolsep{5pt}
    \begin{tabular}{c|cc} \hline
         \textbf{Category} & \textbf{Spring 21} & \textbf{Spring 22}\\
        \hline
        $High_{pre}$ $\rightarrow$ $High_{post}$ & 9 & 13 \\
        $Medium_{pre}$ $\rightarrow$ $High_{post}$ & 8 & 9 \\
        $Low_{pre}$ $\rightarrow$ $High_{post}$ & 3 & 2 \\
        $Low_{pre}$ $\rightarrow$ $Medium_{post}$ & 4 & 5 \\ \hline
        \textbf{Total} & \textbf{24} & \textbf{29} \\
        \hline
    \end{tabular}
    \label{tab:qlg_stat}
\end{table}

\subsection{Trajectories for DRL policy Induction}
Our two DRL policies are induced using pre-collected training data collected by \emph{tutor} making decisions, and many times, the tutor would produce random yet \emph{reasonable} decisions. All students used the same probability tutor, followed the same general procedure as described above, studied the same training materials, and worked through the same training problems.  It has the same  142 features described above and the same three types of pedagogical decisions \emph{but made by the tutor}. The two DRL policies differ in the reward function.

The CQL policy induction was offline, containing \textbf{2,716} students' interaction logs over 13 semesters of classroom studies (Fall 2015 to Spring 2022). Its reward function is based on the students' Normalized Learning Gain (NLG), which measures their learning gain irrespective of their incoming competence. NLG is defined as $\frac{posttest-pretest}{\sqrt{1-pretest}}$, where 1 is the maximum score for both pre- and post-test. 

The CQL-T policy induction was done offline using pre-collected training data containing \textbf{1,819} students' interaction logs over 7 semesters of classroom studies (Spring 2020 to Spring 2023). Its reward function balances NLG and time spent on the tutor.


\subsection{Experimental Settings}

\renewcommand{\arraystretch}{1.5}
\begin{table*}
\footnotesize
    \centering
    \caption{\centering Comparing EM-EDM to baselines on predicting student pedagogical actions (5-fold Cross-validation). The best methods are in bold, and the overall best is highlighted with *}
    \setlength\tabcolsep{5pt}
    \begin{tabular}{c|c|ccccccc} \hline
        \textbf{Category} & \textbf{Methods} & \textbf{Acc} & \textbf{Rec} & \textbf{Prec} & \textbf{F1} & \textbf{AUC} & \textbf{APR} & \textbf{Jaccard} \\
        \hline
        \multirow{2}{*}{\textbf{DRL}} & CQL-T &  0.296(0.05) & 0.274(0.07) & 0.369(0.06) & 0.259(0.03) & 0.454(0.06) & 0.357(0.03) & 0.161(0.03)\\ 
        & \textbf{CQL} &  \textbf{0.376(0.02)} & \textbf{0.358(0.04)} & \textbf{0.329(0.04)} & \textbf{0.325(0.03)} & \textbf{0.526(0.02)} & \textbf{0.382(0.01)} & \textbf{0.221(0.02)} \\ \hline
        \multirow{5}{*}{\textbf{AL}} & BC  &  0.354(0.02) & 0.342(0.02) & 0.346(0.02) & 0.333(0.02) & 0.508(0.02) & 0.339(0.01) & 0.204(0.01)\\ 
        & GAIL &  0.387(0.05) & 0.355(0.03) & 0.360(0.07) & 0.311(0.04) & 0.515(0.02) & 0.343(0.01) & 0.197(0.03)  \\ 
        & AIRL &  0.365(0.03) & 0.375(0.02) & 0.382(0.02) & 0.330(0.02) & 0.533(0.01) & 0.353(0.01) & 0.204(0.02) \\ 
        & \textbf{EDM} &  \textbf{0.737(0.05)} & \textbf{0.708(0.05)} & \textbf{0.723(0.06)} & 0.\textbf{705(0.06)} & \textbf{0.836(0.04)} & \textbf{0.700(0.05)} & \textbf{0.563(0.06)}\\ \cline{2-9}
        & \textbf{EM-EDM} &  \textbf{0.795(0.05)}* & \textbf{0.774(0.06)}* & \textbf{0.776(0.05)}* & \textbf{0.769(0.06)}* & \textbf{0.877(0.02)}* & \textbf{0.762(0.05)}* & \textbf{0.639(0.08)}* \\ \hline
    \end{tabular}
    \label{tab:metrics_cv}
\end{table*}

To evaluate the effectiveness of EM-EDM, we compared it against four AL and two DRL baselines, which include: 

\noindent $\bullet$ \textbf{CQL} \cite{kumar2020conservative}, 
While DQN is one of the most widely applied DRL methods, it can suffer from overestimation bias. But Conservative Q-Learning (CQL) \cite{kumar2020conservative} is designed to address this overestimation bias in Q-learning. This is why we chose CQL as a DRL baseline.

\noindent $\bullet$ \textbf{CQL-T} \cite{kumar2020conservative}, which has the same method as CQL except utilizing a reward function that balances performance and time on task. So when two students learn the same, those who spend less time would get higher rewards.

\noindent $\bullet$ \textbf{BC} \cite{syed2012imitation,raza2012teaching,bain1999bc}, which directly learns a policy using supervised learning on state-action pairs from expert demonstrations. It is a simple approach to learning a policy, but the policy often generalizes poorly and does not recover well from errors.

\noindent $\bullet$ \textbf{GAIL} \cite{ho2016gail}, which is a model-free AL algorithm that learns a policy by simultaneously training it with a discriminator that aims to distinguish expert trajectories against trajectories from the learned policy.

\noindent $\bullet$ \textbf{AIRL} \cite{fu2018learning}, similar to GAIL, adversarially trains a policy against a discriminator that aims to distinguish the expert demonstrations from the learned policy. Unlike GAIL, AIRL recovers a reward function more generalizable to changes in environment dynamics.

\noindent $\bullet$ \textbf{EDM} \cite{jarrett2020strictly}, the state-of-the-art offline AL. It has been demonstrated that EDM can outperform many competitive cutting-edge AL methods with a \emph{single} reward function, thus we will not repetitively conduct all those comparisons here.

\noindent $\bullet$ \textbf{EM-EDM} \cite{xi2023themes}, our proposed method, which assumes the demonstrations follow \emph{multiple} reward functions varying across trajectories (while remaining the same within each trajectory).

We compare the effectiveness of the policies induced by our proposed EM-EDM against four AL-based baselines using the 53 demonstrations and two policies induced by DRL using 2,716 and 1,819 student-ITS interactive trajectories, respectively. All the methods were trained and tested on student interaction trajectories through 5-fold cross-validation; as we do not have access to ground truth rewards, we evaluate performance according to action-matching on held-out test data. We used the same held-out test data to evaluate the performance of CQL and CQL-T-induced policies. We employed the metrics of Accuracy (Acc), Recall (Rec), Precision (Prec), F1-score (F1), AUC, APR, and Jaccard score for evaluation. All the model parameters were determined by cross-validation. For EM-EDM, the optimal cluster number was determined heuristically as $4$ for Task 1 and as $5$ for Task 2 by iteratively implementing the EM until empty clusters were generated or the log-likelihood of the clustering results varied smaller than a pre-defined threshold. Based on our observation, the clustering likelihood for EM-EDM converges within $80$ iterations. For fair comparisons, the optimal parameters in other baselines and ablations were also determined by cross-validation.

\renewcommand{\arraystretch}{1.5}
\begin{table*}
\footnotesize
    \centering
    \caption{\centering Comparing EM-EDM to baselines on predicting future students' pedagogical actions. The best methods are in bold, and the overall best is highlighted with *}
    \setlength\tabcolsep{5pt}
    \begin{tabular}{c|c|ccccccc} \hline
        \textbf{Category} & \textbf{Methods} & \textbf{Acc} & \textbf{Rec} & \textbf{Prec} & \textbf{F1} & \textbf{AUC} & \textbf{APR} & \textbf{Jaccard} \\
        \hline
        \multirow{2}{*}{\textbf{DRL}} & CQL-T  & 0.217 & 0.203 & 0.313 & 0.199 & 0.391 & 0.332 & 0.115 \\ 
        & \textbf{CQL} & \textbf{0.321} & \textbf{0.354} & \textbf{0.284} & \textbf{0.308} & \textbf{0.506} & \textbf{0.366} & \textbf{0.198} \\ \hline
        \multirow{5}{*}{\textbf{AL}} & BC  & 0.328 & 0.334 & 0.344 & 0.317 & 0.501 & 0.335 & 0.190 \\ 
        & GAIL & 0.393 & 0.362 & 0.387 & 0.337 & 0.522 & 0.345 & 0.210 \\ 
        & AIRL & 0.341 & 0.355 & 0.351 & 0.331 & 0.517 & 0.343 & 0.202 \\ 
        & \textbf{EDM} & \textbf{0.669} & \textbf{0.654} & \textbf{0.663} & \textbf{0.656} & \textbf{0.789} & \textbf{0.656} & \textbf{0.498} \\ \cline{2-9}   
        & \textbf{EM-EDM} &  \textbf{0.745}* & \textbf{0.740}* & \textbf{0.745}* & \textbf{0.740}* & \textbf{0.871}* & \textbf{0.765}* & \textbf{0.589}* \\ \hline
    \end{tabular}
    \label{tab:metrics_xsem}
\end{table*}

\subsection{Evaluation Metrics}
We evaluate and compare performance using Accuracy, Precision, Recall, F1 Score, AUC (Area Under the ROC Curve), APR (Area Under the Precision-Recall Curve), and Jaccard score. Jaccard score is used to assess the similarity between the predicted set and the true set of labels, calculated by the size of the intersection divided by the size of the union of two label sets. Because of our task's nature, we consider AUC and Jaccard Score the most crucial metrics, as they are thought to be generally robust.

\section{Results}
We compare our proposed EM-EDM against the six baselines (two DRL-based and four AL-based) on predicting student pedagogical actions. More specifically, they are compared on two related but different tasks. In Task 1, we follow the standard AL research by training and evaluating all models on all the 53 good trajectories through 5-fold Cross-validation \cite{jarrett2020strictly}. Task 2 is more relevant and practical to the educational data mining field by using Spring 21's good trajectories (24) as training and Spring 22's good trajectories (29) trajectories as testing. Our goal in Task 2 is to check whether our clustering results trained from the previous semester can be used to predict future semester students' actions. We also investigate the distribution of the clusters learned from both tasks and check for significant differences among them.

\subsection{Modeling Heterogeneous Student Pedagogical Strategies}

\subsubsection{Comparing Seven Approaches on Student Pedagogical Action Prediction}
Table \ref{tab:metrics_cv} reports our results from Task 1, comparing the seven approaches to student pedagogical action prediction using all 53 good trajectories by 5-fold cross-validation. The best results among baselines are in bold, and the overall best results are highlighted with *. 

\begin{figure}
  \centering

  \begin{subfigure}{0.5\textwidth}
    \centering
    \includegraphics[width=\linewidth]{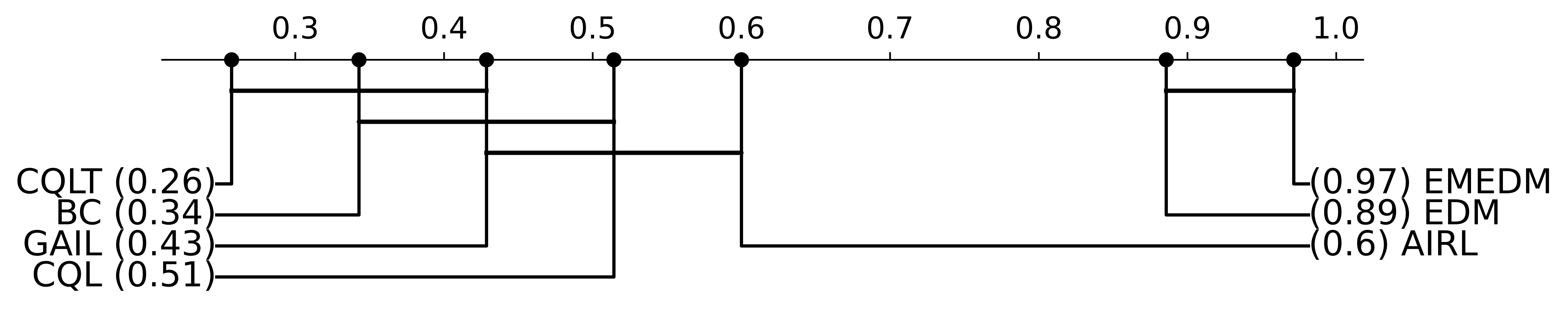}
    \caption{AUC}
    \label{subfig:cd_auc}
    \Description{AUC}
  \end{subfigure}%
  \hfill
  \begin{subfigure}{0.5\textwidth}
    \centering
    \includegraphics[width=\linewidth]{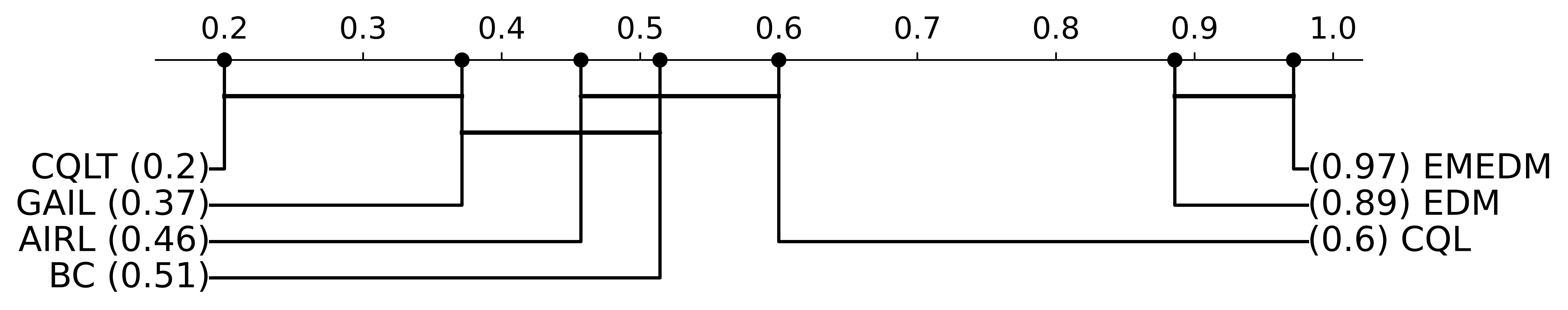}
    \caption{Jaccard Score}
    \label{subfig:cd_jac}
    \Description{Jaccard Score}
  \end{subfigure}

  \caption{Critical difference diagram with Conover Friedman test over (a) AUC and (b) Jaccard Score.}
  \label{fig:cd_plot}
\end{figure}

To better illustrate our results, we use the Critical Difference (CD) diagram, a powerful tool to statistically compare several classifiers' performance on multiple datasets in a robust way \cite{demvsar2006statistical}. At first, we conducted the Friedman test to check if there is any significant difference among the models across the folds of cross-validation in terms of AUC and Jaccard scores. If there is a significant difference, then we conduct the pairwise Conover post hoc test \cite{friedman1937rank}, \cite{conover1979comparison} to get the ranks and plot the CD diagrams (Figure \ref{fig:cd_plot}). In the CD diagrams in Figure \ref{fig:cd_plot}, the position of the models represents their mean ranks according to the corresponding metric across all folds, where low ranks indicate that a model wins more often than its competitors with higher ranks. Two or more models are connected with each other if we can not tell their performances apart, in the sense of statistical significance, with a confidence level of $0.05$.

As shown in Table \ref{tab:metrics_cv}, EM-EDM outperforms all the baselines across all evaluation metrics and the pairwise differences in terms of AUC and Jaccard score are significant according to the CD diagram in Figure \ref{fig:cd_plot}. EDM outperforms all other baselines except EM-EDM assuming a \emph{single} reward function across the trajectories which proves it's strength in inducing policy from demonstrations as a state-of-the-art offline AL method. But when EM-EDM outperforms EDM, it indicates that there is heterogeneity among the 53 students' pedagogical strategies and considering the heterogeneity would be a better way for modeling complex human-centric tasks, such as student strategy recognition in education. Besides, probably the reason behind the other AL methods did not perform as good as the EDM, is because they cannot take the state distribution into account when learning the policy. 

However, among the DRL baselines, CQL performs better than CQL-T, which suggests that the students are more likely to be learning efficiency-oriented rather than time efficiency-oriented as the reward function of CQL-T has time in it while CQL does not. 

Though among the DRL methods, CQL performs better, it falls short when compared to EDM and EM-EDM even after being trained with a large amount of data, which shows us that EM-EDM is a much more data-efficient approach for modeling student pedagogical strategies.

\subsubsection{Distribution of the Discovered Clusters}

When applied to Task 1, EM-EDM discovered 4 clusters or subgroups among all the 53 good student trajectories. The number of students per subgroup are 25(47.17\%), 13(24.53\%), 11(20.75\%) and 4(7.55\%) which shows an imbalance distribution. A one-way ANOVA test showed no significant difference in the subgroups' learning performances like post-test score, NLG, Iso\_NLG, and Training Time. First, since we used only the good trajectories, it is likely that the students would not have much difference in their learning performances.

\subsection{Future Student's Pedagogical Action Prediction leveraging the Past}
\subsubsection{Clustering Prediction Tasks}
Table \ref{tab:metrics_xsem} reports our findings from semester-based evaluation where we trained the models on Spring 21 semester's 24 good trajectories to learn the clusters and using them to predict Spring 22 semester's 29 good trajectories pedagogical actions. The best results among baselines are in bold, and the overall best results are highlighted with *. 

As shown in Table \ref{tab:metrics_xsem}, EM-EDM outperforms all baselines across all evaluation metrics in predicting future students' pedagogical actions too. The pairwise performance comparison outcomes among the seven methods are similar to Task 1. However, it is important to emphasize that this task of future students' pedagogical action prediction using a past one is a much stricter approach than the standard cross-validation \cite{mao2021dissertation}. This result indicates that we can model student strategies over time despite population change, probably because there are strategies that have followers from both populations.

\subsubsection{Distribution among the Discovered Clusters}

When applied on Task 2, EM-EDM discovered 5 clusters or subgroups among Spring 21 semester's 24 good trajectories. Using the clustering results obtained, Spring 22 semester's 29 good trajectories were clustered. Table \ref{tab:dist_task2} shows both the semesters' distribution of students across those 5 clusters. We found no significant difference between two semester's cluster distributions, $\chi^2(1, 5) = 3.038, p = .551$.

\renewcommand{\arraystretch}{1.5}
\begin{table}
\footnotesize
    \centering
    \caption{\centering Distribution among the Discovered Clusters in Task 2 (C0-C4 refers to the 5 clusters)}
    \setlength\tabcolsep{5pt}
    \begin{tabular}{c|ccccc|c} 
        \hline
        Semester & C0 & C1 & C2 & C3 & C4 & Total\\\hline
        Spring 21 & 3 & 9 & 8 & 2 & 2 & 24\\
        Spring 22 & 6 & 5 & 11 & 3 & 4 & 29\\
        \hline
    \end{tabular}
    \label{tab:dist_task2}
\end{table}

\section{Discussions}
Results suggest that our proposed EM-EDM framework outperforms all the chosen baselines on both tasks for students' next pedagogical action prediction. The baselines we utilized are considered state-of-the-art and sufficiently fair for the prediction task at hand. In a related paper \cite{jarrett2020strictly}, BC was employed as the initial baseline for a similar prediction task involving the real-world healthcare dataset MIMIC-III. Notably, BC outperformed other baselines in 2-Action settings based on AUC metrics. However, GAIL and AIRL exhibit superior capabilities compared to BC, as they employ adversarial training techniques against a discriminator tasked with distinguishing expert demonstrations from learned policies, thus justifying their selection as better baselines \cite{ho2016gail, fu2018learning}. We posit that the under-performance of the baselines may be attributed to the irregular and noisy nature of the student-interaction log data, which is inherently challenging to model. Nevertheless, EDM and EM-EDM demonstrate proficiency in handling such complexities owing to their strengths in energy-based modeling and consideration of heterogeneity. Moreover, our proposed EM-EDM framework is built upon the premise of recognizing the heterogeneity in student pedagogical decision-making strategies. Clustering allows us to group students who follow similar strategies together, thereby enabling cluster-based statistical analyses. As detailed in Sections 5.1.2 and 5.2.2 of our paper, these analyses provide additional insights that can inform future research directions.

\section{Conclusions}
In this study, we investigated heterogeneous student pedagogical strategy modeling via our proposed EM-EDM, a general AL framework, which induces effective pedagogical policies from given optimal or near-optimal demonstrations. To evaluate the performance of EM-EDM, we apply it on two different but related tasks which involves pedagogical action prediction and compare against four AL-based baselines and two policies induced by DRL. The results showed that, for both tasks, EM-EDM outperforms the four AL baselines across all performance metrics as well as the two DRL baselines. We believe, the key impact of this study lies in the capability of EM-EDM to effectively model complex student pedagogical decision-making process through the ability to manage a large, continuous state space and its adaptability in handling diverse and heterogeneous reward functions. The results obtained from EM-EDM showed that students' demonstrations exhibit heterogeneous reward functions. Moreover, EM-EDM can induce distinct pedagogical policies for different reward functions even with as few as 24 demonstrations. In the future, we will carry out this study for upcoming semesters. We will also investigate evolving reward function across demonstrations.


\section{Acknowledgments}
This material is based upon work supported by the National Science Foundation AI Institute for Engaged Learning (EngageAI Institute) under Grant No. DRL-2112635, Generalizing Data-Driven Technologies to Improve Individualized STEM Instruction by Intelligent Tutors (2013502), Integrated Data-driven Technologies for Individualized Instruction in STEM Learning Environments (1726550),  CAREER: Improving Adaptive Decision Making in Interactive Learning Environments (1651909). Any opinions, findings, and conclusions expressed in this material are those of the authors and do not necessarily reflect the views of the NSF.
%
\bibliographystyle{abbrv}
\bibliography{edm24}  

\begin{thebibliography}{10}

\bibitem{abbeel2004apprenticeship}
P.~Abbeel and A.~Y. Ng.
\newblock Apprenticeship learning via inverse reinforcement learning.
\newblock In {\em Proceedings of the 21st international conference on Machine learning}, page~1. ACM, 2004.

\bibitem{abdelshiheed2023bridging}
M.~Abdelshiheed, J.~W. Hostetter, T.~Barnes, and M.~Chi.
\newblock Bridging declarative, procedural, and conditional metacognitive knowledge gap using deep reinforcement learning.
\newblock In {\em Proceedings of the Annual Meeting of the Cognitive Science Society}, number~45 in CogSci '23', 2023.

\bibitem{abdelshiheed2023leveraging}
M.~Abdelshiheed, J.~W. Hostetter, T.~Barnes, and M.~Chi.
\newblock Leveraging deep reinforcement learning for metacognitive interventions across intelligent tutoring systems.
\newblock In {\em International Conference on Artificial Intelligence in Education}, pages 291--303. Springer, 2023.

\bibitem{amodei2016concrete}
D.~Amodei, C.~Olah, J.~Steinhardt, P.~Christiano, J.~Schulman, and D.~Man{\'e}.
\newblock Concrete problems in ai safety.
\newblock {\em arXiv preprint arXiv:1606.06565}, 2016.

\bibitem{arora2021min}
S.~Arora, P.~Doshi, and B.~Banerjee.
\newblock Min-max entropy inverse rl of multiple tasks.
\newblock In {\em IEEE International Conference on Robotics and Automation (ICRA)}, pages 12639--12645, 2021.

\bibitem{asoh2013application}
H.~Asoh, M.~S.~S. Akaho, T.~Kamishima, K.~Hasida, E.~Aramaki, and T.~Kohro.
\newblock An application of inverse reinforcement learning to medical records of diabetes treatment.
\newblock In {\em ECMLPKDD2013 workshop on reinforcement learning with generalized feedback}, 2013.

\bibitem{DBLP:conf/aied/AusinMBC20}
M.~S. Ausin, M.~Maniktala, T.~Barnes, and M.~Chi.
\newblock Exploring the impact of simple explanations and agency on batch deep reinforcement learning induced pedagogical policies.
\newblock In {\em AIED}, pages 472--485, 2020.

\bibitem{babes2011apprenticeship}
M.~Babes, V.~Marivate, K.~Subramanian, and M.~L. Littman.
\newblock Apprenticeship learning about multiple intentions.
\newblock In {\em Proceedings of the 28th ICML}, pages 897--904, 2011.

\bibitem{bain1999bc}
M.~Bain and C.~Sommut.
\newblock A framework for behavioural cloning.
\newblock In {\em Machine Intelligence}, 1999.

\bibitem{chi2011empirically}
M.~Chi, K.~VanLehn, D.~Litman, and P.~Jordan.
\newblock Empirically evaluating the application of reinforcement learning to the induction of effective and adaptive pedagogical strategies.
\newblock {\em User Modeling and User-Adapted Interaction}, 21(1-2):137--180, 2011.

\bibitem{chi2011evaluation}
M.~Chi, K.~VanLehn, D.~Litman, and P.~Jordan.
\newblock An evaluation of pedagogical tutorial tactics for a natural language tutoring system: A reinforcement learning approach.
\newblock {\em International Journal of Artificial Intelligence in Education}, 21(1-2):83--113, 2011.

\bibitem{choi2012nonparametric}
J.~Choi and K.-E. Kim.
\newblock Nonparametric bayesian inverse reinforcement learning for multiple reward functions.
\newblock In {\em Advances in Neural Information Processing Systems}, pages 305--313, 2012.

\bibitem{conover1979comparison}
W.~J. Conover and R.~L. Iman.
\newblock Multiple-comparisons procedures. informal report.
\newblock {\em Technical Report}, 0(0), 2 1979.

\bibitem{demvsar2006statistical}
J.~Dem{\v{s}}ar.
\newblock Statistical comparisons of classifiers over multiple data sets.
\newblock {\em The Journal of Machine learning research}, 7:1--30, 2006.

\bibitem{dimitrakakis2011bayesian}
C.~Dimitrakakis and C.~A. Rothkopf.
\newblock Bayesian multitask inverse reinforcement learning.
\newblock In {\em European workshop on reinforcement learning}, pages 273--284. Springer, 2011.

\bibitem{finn2016guided}
C.~Finn, S.~Levine, and P.~Abbeel.
\newblock Guided cost learning: Deep inverse optimal control via policy optimization.
\newblock In {\em International conference on machine learning}, pages 49--58. PMLR, 2016.

\bibitem{friedman1937rank}
M.~Friedman.
\newblock The use of ranks to avoid the assumption of normality implicit in the analysis of variance.
\newblock {\em Journal of the American Statistical Association}, 32(200):675--701, 1937.

\bibitem{fu2018learning}
J.~Fu, K.~Luo, and S.~Levine.
\newblock Learning robust rewards with adversarial inverse reinforcement learning, 2018.

\bibitem{gonzalez2007innovative}
W.~J. Gonz{\'a}lez-Espada and D.~W. Bullock.
\newblock Innovative applications of classroom response systems: Investigating students’ item response times in relation to final course grade, gender, general point average, and high school act scores.
\newblock {\em Electronic Journal for the Integration of Technology in Education}, 6:97--108, 2007.

\bibitem{goodfellow2014generative}
I.~J. Goodfellow, J.~Pouget-Abadie, M.~Mirza, B.~Xu, D.~Warde-Farley, S.~Ozair, A.~Courville, and Y.~Bengio.
\newblock Generative adversarial networks, 2014.

\bibitem{goyal2019using}
P.~Goyal, S.~Niekum, and R.~J. Mooney.
\newblock Using natural language for reward shaping in reinforcement learning.
\newblock {\em arXiv preprint arXiv:1903.02020}, 2019.

\bibitem{grathwohl2019your}
W.~Grathwohl, K.-C. Wang, J.-H. Jacobsen, D.~Duvenaud, M.~Norouzi, and K.~Swersky.
\newblock Your classifier is secretly an energy based model and you should treat it like one.
\newblock {\em arXiv preprint arXiv:1912.03263}, 2019.

\bibitem{ho2016generative}
J.~Ho and S.~Ermon.
\newblock Generative adversarial imitation learning.
\newblock {\em Advances in neural information processing systems}, 29, 2016.

\bibitem{ho2016gail}
J.~Ho and S.~Ermon.
\newblock Generative adversarial imitation learning.
\newblock In D.~Lee, M.~Sugiyama, U.~Luxburg, I.~Guyon, and R.~Garnett, editors, {\em Advances in Neural Information Processing Systems}, volume~29. Curran Associates, Inc., 2016.

\bibitem{hostetter2023leveraging}
J.~W. Hostetter, M.~Abdelshiheed, T.~Barnes, and M.~Chi.
\newblock Leveraging fuzzy logic towards more explainable reinforcement learning-induced pedagogical policies on intelligent tutoring systems.
\newblock In {\em 2023 {IEEE} International Conference on Fuzzy Systems}. IEEE, 2023.

\bibitem{hostetter2023self}
J.~W. Hostetter, M.~Abdelshiheed, T.~Barnes, and M.~Chi.
\newblock A self-organizing neuro-fuzzy q-network: Systematic design with offline hybrid learning.
\newblock In {\em Proceedings of the 22nd International Conference on Autonomous Agents and Multiagent Systems (AAMAS)}, 2023.

\bibitem{hostetter2023xai}
J.~W. Hostetter, C.~Conati, X.~Yang, M.~Abdelshiheed, T.~Barnes, and M.~Chi.
\newblock Xai to increase the effectiveness of an intelligent pedagogical agent.
\newblock In {\em Proceedings of the 23rd ACM International Conference on Intelligent Virtual Agents}, IVA '23, New York, NY, USA, 2023. Association for Computing Machinery.

\bibitem{iglesias2009learning}
A.~Iglesias, P.~Mart{\'\i}nez, R.~Aler, and F.~Fern{\'a}ndez.
\newblock Learning teaching strategies in an adaptive and intelligent educational system through reinforcement learning.
\newblock {\em Applied Intelligence}, 31(1):89--106, 2009.

\bibitem{jarrett2020strictly}
D.~Jarrett, I.~Bica, and M.~van~der Schaar.
\newblock Strictly batch imitation learning by energy-based distribution matching.
\newblock {\em Advances in Neural Information Processing Systems}, 33:7354--7365, 2020.

\bibitem{ju2022student}
S.~Ju, X.~Yang, T.~Barnes, and M.~Chi.
\newblock Student-tutor mixed-initiative decision-making supported by deep reinforcement learning.
\newblock In {\em International Conference on Artificial Intelligence in Education}, pages 440--452. Springer, 2022.

\bibitem{ju2020aiedcritical}
S.~Ju, G.~Zhou, M.~Abdelshiheed, T.~Barnes, and M.~Chi:.
\newblock Evaluating critical reinforcement learning framework in the field.
\newblock {\em AIED}, pages 215--227, 2021.

\bibitem{kaelbling1996reinforcement}
L.~P. Kaelbling, M.~L. Littman, and A.~W. Moore.
\newblock Reinforcement learning: A survey.
\newblock {\em Journal of artificial intelligence research}, 4:237--285, 1996.

\bibitem{kostrikov2018discriminator}
I.~Kostrikov, K.~K. Agrawal, D.~Dwibedi, S.~Levine, and J.~Tompson.
\newblock Discriminator-actor-critic: Addressing sample inefficiency and reward bias in adversarial imitation learning.
\newblock {\em International Conference on Learning Representations}, 2018.

\bibitem{kostrikov2019imitation}
I.~Kostrikov, O.~Nachum, and J.~Tompson.
\newblock Imitation learning via off-policy distribution matching.
\newblock {\em International Conference on Learning Representations}, 2020.

\bibitem{kumar2020conservative}
A.~Kumar, A.~Zhou, G.~Tucker, and S.~Levine.
\newblock Conservative q-learning for offline reinforcement learning.
\newblock {\em Advances in Neural Information Processing Systems}, 33:1179--1191, 2020.

\bibitem{lin2017comparisons}
C.~Lin and M.~Chi.
\newblock A comparisons of bkt, rnn and lstm for learning gain prediction.
\newblock In E.~Andr{\'e}, R.~Baker, X.~Hu, M.~M.~T. Rodrigo, and B.~du~Boulay, editors, {\em Artificial Intelligence in Education}, pages 536--539. Springer, 2017.

\bibitem{mandel2014offline}
T.~Mandel et~al.
\newblock Offline policy evaluation across representations with applications to educational games.
\newblock In {\em AAMAS}, pages 1077--1084, 2014.

\bibitem{mao2021dissertation}
Y.~Mao.
\newblock Temporal skill discovery for modeling student learning progression across stem domains.
\newblock {\em PhD thesis}, 2001.

\bibitem{mao2018tracing}
Y.~Mao, C.~Lin, and M.~Chi.
\newblock Deep learning vs. bayesian knowledge tracing: Student models for interventions.
\newblock {\em Journal of Educational Data Mining}, 2018.

\bibitem{najar2014}
A.~S. Najar, A.~Mitrovic, and B.~M. McLaren.
\newblock Adaptive support versus alternating worked examples and tutored problems: Which leads to better learning?
\newblock In {\em UMAP}, pages 171--182. Springer, 2014.

\bibitem{pan2019dissecting}
M.~Pan, Y.~Li, X.~Zhou, Z.~Liu, R.~Song, H.~Lu, and J.~Luo.
\newblock Dissecting the learning curve of taxi drivers: A data-driven approach.
\newblock In {\em Proceedings of the 2019 SIAM International Conference on Data Mining}, pages 783--791. SIAM, 2019.

\bibitem{pomerleau1991efficient}
D.~A. Pomerleau.
\newblock Efficient training of artificial neural networks for autonomous navigation.
\newblock {\em Neural computation}, 3(1):88--97, 1991.

\bibitem{rafferty2015inferring}
A.~N. Rafferty, M.~M. LaMar, and T.~L. Griffiths.
\newblock Inferring learners' knowledge from their actions.
\newblock {\em Cognitive Science}, 39(3):584--618, 2015.

\bibitem{raza2012teaching}
S.~Raza, S.~Haider, and M.-A. Williams.
\newblock Teaching coordinated strategies to soccer robots via imitation.
\newblock In {\em IEEE International Conference on Robotics and Biomimetics}, pages 1434--1439. IEEE, 2012.

\bibitem{renkl2002}
A.~Renkl, R.~K. Atkinson, U.~H. Maier, and R.~Staley.
\newblock From example study to problem solving: Smooth transitions help learning.
\newblock {\em J Exp Educ}, 70(4):293--315, 2002.

\bibitem{rowe2015improving}
J.~P. Rowe and J.~C. Lester.
\newblock Improving student problem solving in narrative-centered learning environments: A modular reinforcement learning framework.
\newblock In {\em AIED}, pages 419--428. Springer, 2015.

\bibitem{schnipke2002exploring}
D.~L. Schnipke and D.~J. Scrams.
\newblock Exploring issues of examinee behavior: Insights gained from response-time analyses.
\newblock {\em Computer-based testing: Building the foundation for future assessments}, pages 237--266, 2002.

\bibitem{shen2016reinforcement}
S.~Shen and M.~Chi.
\newblock Reinforcement learning: the sooner the better, or the later the better?
\newblock In {\em UMAP}, pages 37--44. ACM, 2016.

\bibitem{syed2012imitation}
U.~Syed and R.~E. Schapire.
\newblock Imitation learning with a value-based prior.
\newblock {\em arXiv preprint arXiv:1206.5290}, 2012.

\bibitem{van2011effects}
T.~Van~Gog, L.~Kester, and F.~Paas.
\newblock Effects of worked examples, example-problem, and problem-example pairs on novices’ learning.
\newblock {\em Contemporary Educational Psychology}, 36(3):212--218, 2011.

\bibitem{wang2020inferring}
J.~Wang, D.~Roberts, and A.~Enquobahrie.
\newblock Inferring continuous treatment doses from historical data via model-based entropy-regularized reinforcement learning.
\newblock In {\em Asian Conference on Machine Learning}, pages 433--448. PMLR, 2020.

\bibitem{wang2017interactive}
P.~Wang, J.~Rowe, W.~Min, B.~Mott, and J.~Lester.
\newblock Interactive narrative personalization with deep reinforcement learning.
\newblock In {\em IJCAI}, 2017.

\bibitem{welling2011bayesian}
M.~Welling and Y.~W. Teh.
\newblock Bayesian learning via stochastic gradient langevin dynamics.
\newblock In {\em Proceedings of the 28th international conference on machine learning (ICML-11)}, pages 681--688, 2011.

\bibitem{xi2023themes}
X.~Yang, G.~Gao, and M.~Chi.
\newblock An offline time-aware apprenticeship learning framework for evolving reward functions.
\newblock {\em arXiv preprint arxiv:2305.09070v1}, 2023.

\bibitem{yang2020student}
X.~Yang, G.~Zhou, M.~Taub, R.~Azevedo, and M.~Chi.
\newblock Student subtyping via em-inverse reinforcement learning.
\newblock {\em International Educational Data Mining Society}, 2020.

\bibitem{guojing2017tutorstudent}
G.~Zhou and M.~Chi.
\newblock The impact of decision agency \& granularity on aptitude treatment interaction in tutoring.
\newblock {\em CogSci}, pages 3652--3657, 2017.

\bibitem{ziebart2008maximum}
B.~D. Ziebart, A.~L. Maas, J.~A. Bagnell, A.~K. Dey, et~al.
\newblock Maximum entropy inverse reinforcement learning.
\newblock In {\em AAAI}, volume~8, pages 1433--1438. Chicago, IL, USA, 2008.

\end{thebibliography}
\balancecolumns
\end{document}